\pdfoutput=1
\documentclass[11pt]{article}
\usepackage{ACL2023}
\usepackage{times}
\usepackage[T1]{fontenc}
\usepackage[utf8]{inputenc}
\usepackage{microtype}
\usepackage{amsmath}
\usepackage{braket}
\usepackage{subfigure}
\usepackage[export]{adjustbox}
\usepackage{url}
\usepackage{bm}
\usepackage{tikz}
\usepackage[ruled,vlined,algo2e]{algorithm2e}
\usepackage{algorithm}
\usepackage{algorithmic}
\usepackage{setspace}
\usepackage{hyperref}
\usepackage{enumitem}
\usepackage{makecell}
\usepackage{amssymb}
\usepackage{pifont}

\newcommand{\xmark}{\ding{55}}
\usepackage{tabularx}
\let\Algorithm\algorithm
\renewcommand\algorithm[1][]{\Algorithm[#1]\setstretch{1}}

\usepackage{xcolor}
\title{UniTRec: A Unified Text-to-Text Transformer and Joint Contrastive Learning Framework for Text-based Recommendation}
\author{
  Zhiming Mao$^{1,2}$, Huimin Wang$^{1,3}$, Yiming Du$^{1,2}$, Kam-Fai Wong$^{1,2}$ \\
  $^1$The Chinese University of Hong Kong, Hong Kong, China \\
  $^2$MoE Key Laboratory of High Confidence Software Technologies, China \\
  $^3$Jarvis Lab, Tencent, Shenzhen, China \\
  \texttt{\{zmmao,ydu,kfwong\}@se.cuhk.edu.hk} \\
  \texttt{hmmmwang@tencent.com}
}

\begin{document}
\maketitle
\begin{abstract}
Prior study has shown that pretrained language models~(PLM) can boost the performance of text-based recommendation. In contrast to previous works that either use PLM to encode user history as a whole input text, or impose an additional aggregation network to fuse multi-turn history representations, we propose a unified local- and global-attention Transformer encoder to better model two-level contexts of user history. Moreover, conditioned on user history encoded by Transformer encoders, our framework leverages Transformer decoders to estimate the language perplexity of candidate text items, which can serve as a straightforward yet significant contrastive signal for user-item text matching. Based on this, our framework, UniTRec, unifies the contrastive objectives of discriminative matching scores and candidate text perplexity to jointly enhance text-based recommendation. Extensive evaluation shows that UniTRec delivers SOTA performance on three text-based recommendation tasks.\footnote{Our code is available at \href{https://github.com/Veason-silverbullet/UniTRec}{https://github.com/Veason-silverbullet/UniTRec}.}
\end{abstract}

\section{Introduction}
Text-based recommendation~\citep{Trec1, Trec2, DAE_RNN, RecoBERT} aims to recommend relevant textual content~(e.g., news articles, Twitter posts) to people based on their behaviors as represented in historical log texts. For instance, engagement recommendation~\citep{EngageRec_dataset} on social media~(e.g., Twitter and Reddit) helps users discover and engage with interested threads by modeling their browsing history.

Pretrained language models~\citep{BERT, GPT3} have made waves in recent text-based recommendation research~\citep{UNBERT, Quoter, P5}. The most common practice is using PLM encoders (BERT family) to learn representations of user history and candidate item texts. Recommendation matching scores are computed over the user and item representations and finally optimized by noise contrastive estimation~(NCE) loss \citep{NCE_loss} for ranking multiple candidates.

Unlike encoding single text, using PLM to encode multi-turn texts of user history is nontrivial. Existing works~\citep{RecoBERT, Quoter, P5} concatenate multi-turn history texts as a whole input text, then use one PLM encoder to learn the holistic user representation. This is a standard PLM encoding manner but ignores the relation among history turns, as all word tokens from different history turns are \textit{equally attended}\footnote{There is no inductive bias of turn-level and history-level relations introduced to Transformer self-attention computation, where each token plays an equal role.}. In contrast, previous studies point out that learning the relation among user history turns is also beneficial \citep{DCR, HieRec}. Another approach is using PLM encoders to learn representations from multi-turn history texts, followed by an additional aggregation network to fuse the multi-turn representations~\citep{PLM_newsrec, MINER}. However, the imposed aggregation networks (with newly initialized parameters) weaken the representation power of PLM encoders which are already pretrained on large-scale corpora.

This work introduces UniTRec, a \textbf{Uni}fied text-to-text \textbf{T}ransformer framework for text-based \textbf{Rec}om-mendation. In the encoder component of UniTRec, we design local- and global-attention to learn user history representations through tailored attention masking, which aims to jointly model word-level and turn-level relations of user history. UniTRec can utilize the full power of PLM encoders because it preserves the intact structure of PLM encoders without newly imposed parameters.

\begin{figure}[t]
\centering
\includegraphics[width=79.875mm]{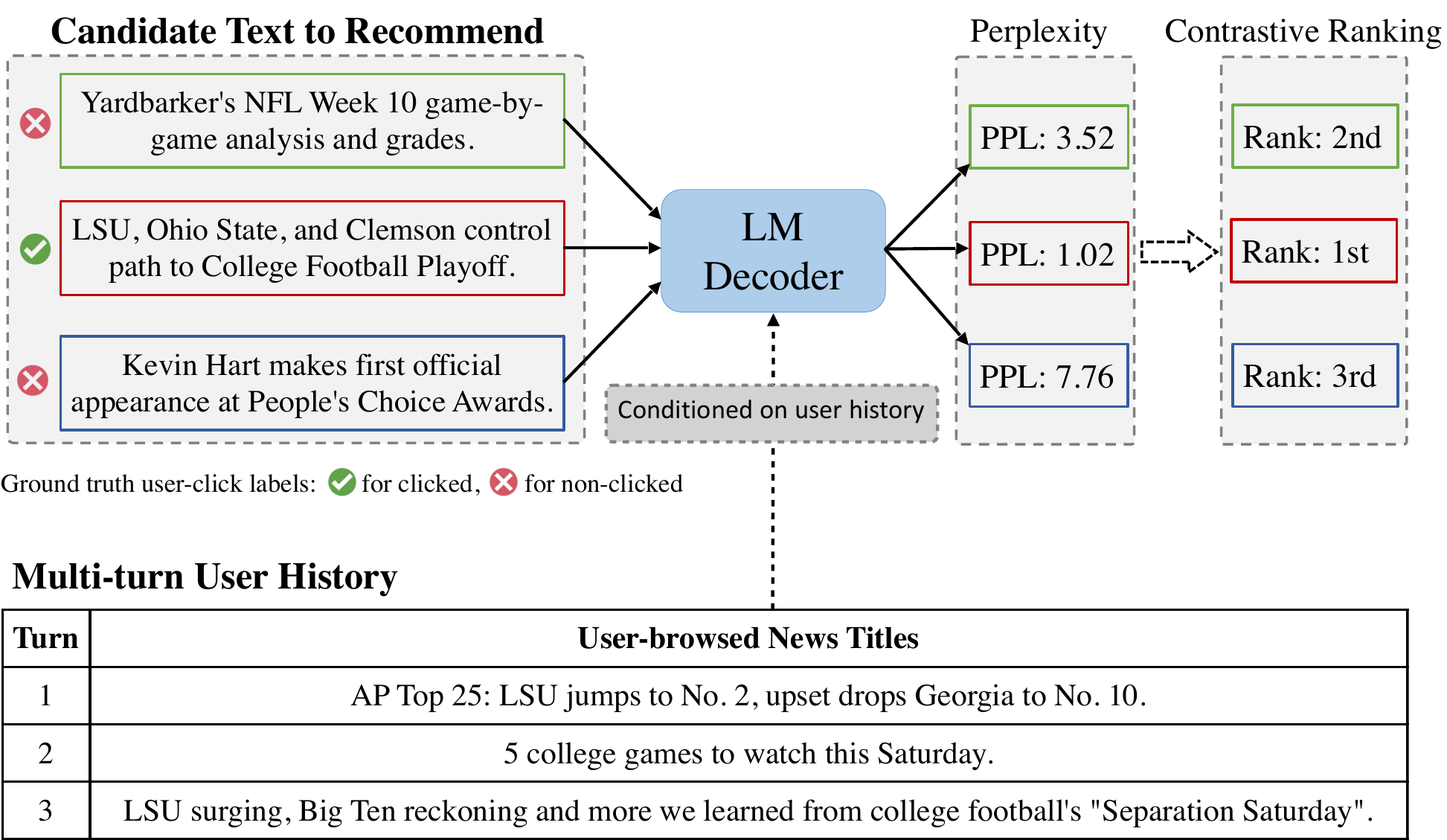}
\captionsetup{font=10pt}
\caption{An example of perplexity-based ranking for candidate item texts, conditioned on user history. The illustrated task is text-based news recommendation.}
\label{fig:example}
\end{figure}
Different from most previous works that predict user-candidate matching scores solely based on the representations learned by Transformer encoders, we argue that conditioned on user representations learned by Transformer encoders, candidate text perplexity (PPL) estimated by pretrained Transformer decoders is also a straightforward yet significant signal for text-based recommendation. As shown in Figure~\ref{fig:example}, we hypothesize that the candidate text perplexity estimated by pretrained LM decoders can directly measure the text matching degree between user history and candidate texts. It is because the perplexity estimates the likelihood of candidate texts based on encoder outputs, which naturally indicates the probabilities of candidate texts given the user history. Besides, UniTRec can use the last hidden states of Transformer decoders to directly predict matching scores. Hence, this work unifies the contrastive objectives of discriminative matching scores and candidate text perplexity to jointly enhance text-based recommendation.

The contributions of this work are: ($1$) We propose local- and global-attention to model two-level relation of user history without additional parameters, which enjoys the full power of PLM encoders. ($2$) We introduce PLM perplexity to measure user-candidate text matching and unify the objectives of discriminative matching scores and candidate text perplexity to enhance text-based recommendation. ($3$) Experiments on three text-based recommendation datasets validate the effectiveness of UniTRec.

\section{Approach}
\subsection{Unified User-history Modeling}
Formally, multi-turn history of a user is represented as $H=[t_{1},t_{2},...,t_{N}]$, and each turn text $t_{i}$ contains $|t_{i}|$ words as $t_{i}=[x_{i}^{1},x_{i}^{2},...,x_{i}^{|t_{i}|}]$. UniTRec aims to unify learning word- and turn-level context representations in one Transformer encoder.

\textbf{Local attention on word-level context.} We first concatenate the multi-turn history texts as the input tokens $X=[x_{1}^{1},x_{1}^{2},...,x_{1}^{|t_{1}|},...,x_{N}^{1},x_{N}^{2},...,x_{N}^{|t_{N}|}]$. Inspired by \citet{UniLM}, we tailor the attention masking in Transformer self-attention to learn the word-level context of each turn. Specifically, we allow word tokens from the same turn to attend to each other, while tokens from different turns are excluded from self-attention computation:
\begin{equation}\label{eq1}
\setlength{\abovedisplayskip}{4pt}
\setlength{\belowdisplayskip}{2pt}
\begin{split}
\mathbf{M}_{i,j} &= \begin{cases}
0,\;\quad\;\mbox{token $x_{i}$ and $x_{j}$ in the same turn} \\
-\infty, \;\mbox{otherwise}
\end{cases} \\
\mathrm{Atte}&\mathrm{ntion}(Q, K, V) = \mathrm{softmax}(\frac{QK^{T}}{\sqrt{d_{k}}}\mathbf{+M})V
\end{split}
\end{equation}
, where $Q, K, V$ are self-attention query, key, and value in \citet{Transformer}, $\mathbf{M}$ is the mask matrix to achieve local-attention inside each turn text. The local self-attention blocks consist of $L_{1}$ layers, by which original PLM encoders can be adapted to learn word-level context representations of turns.

\textbf{Global attention on turn-level context.} Over the local self-attention layers, we leverage global self-attention to model the relation among history turns.  Specifically, tokens from all turns attend to each other in self-attention computation (by setting the mask matrix $\mathbf{M}=\mathbf{0}$). In this way, Transformer encoders can perform global interaction among each token (and turn) to learn turn-level context representations of user history. There are $L_{2}$ layers in the global self-attention blocks, which can also be inherited from PLM encoders directly.

\subsection{Joint Contrastive Ranking Objectives}
Conditioned on the history representation, we input the candidate text to Transformer decoders to predict how likely it should be recommended. It is worth noting that Transformer decoders can naturally perform effective \textbf{cross-attention} interaction between history and candidate hidden states.
\subsubsection{Objective on Discriminative Scores}
Motivated by \citet{BART}, we feed the last hidden state of decoder output $h_{T}$ to an MLP score-head which predicts the user-candidate matching score $S^{d}=\mathrm{ScoreHead}(h_{T})$. The matching score is discriminative, as higher scores indicate higher user-candidate matching probabilities.

\begin{figure*}[t]
\centering
\includegraphics[width=125.06625mm]{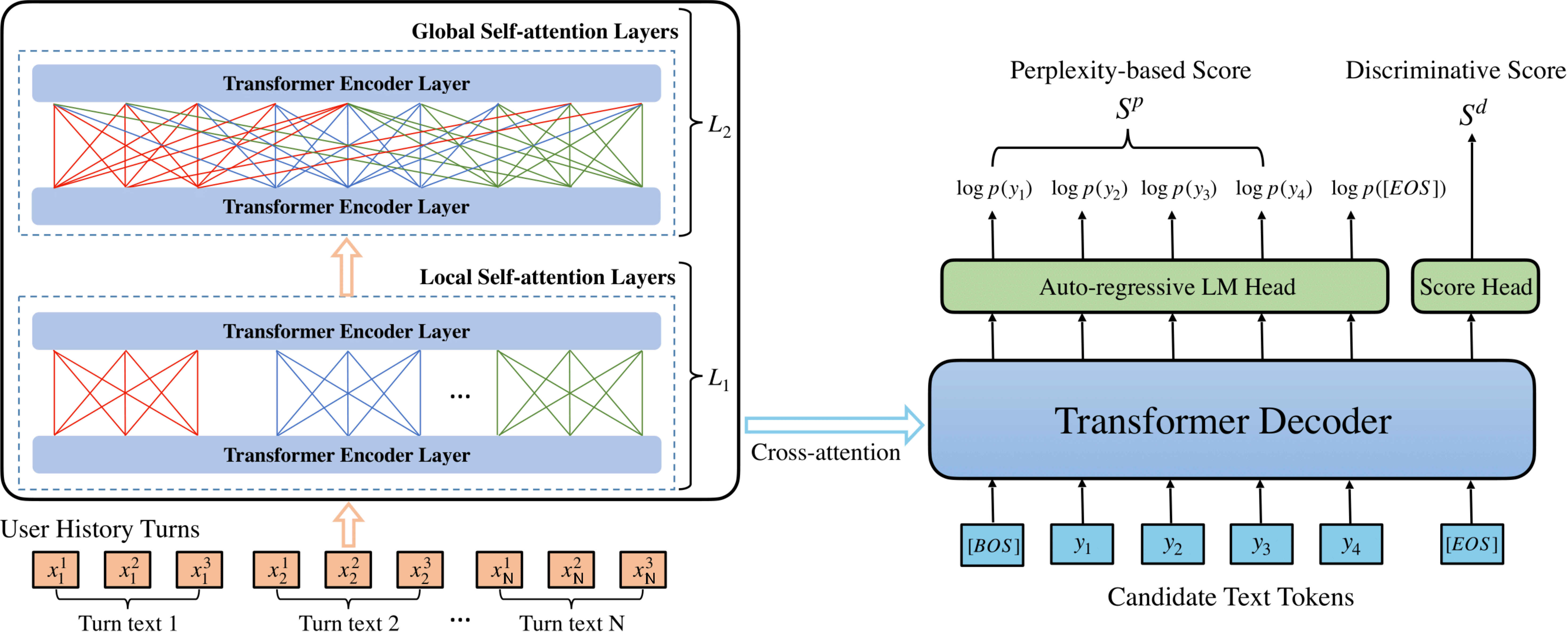}
\captionsetup{font=10pt}
\vskip -0.28mm
\caption{Overview of UniTRec. In training, matching scores $S^{d}$ and $S^{p}$ are optimized by the NCE loss, respectively. In inference, $S^{d}$ and $S^{p}$ are normalized and combined to derive the final output ranking.}
\label{fig:model}
\vskip -0.28mm
\end{figure*}
Following previous works \citep{MINER, Quoter}, we adopt negative sampling with NCE loss to optimize matching score prediction. Given the user history and its ground truth matched candidate $C_{i}$, UniTRec predicts the matching score as $S^{d+}_i$. In addition, $K$ unmatched negative candidates $\{C_{j}\}_{j=1}^{K}$ are sampled from the candidate set, and their matching scores are $\{S^{d-}_j\}_{j=1}^{K}$. The NCE loss is represented in a contrastive form:
\begin{equation}\label{eq2}
\setlength{\abovedisplayskip}{4pt}
\setlength{\belowdisplayskip}{4pt}
\mathcal{L}_{i}^{d} = -\log\frac{\mathrm{exp}(S^{d+}_i)}{\mathrm{exp}(S^{d+}_i)+\sum\nolimits_{j=1}^{K}{\mathrm{exp}(S^{d-}_{j})}}
\end{equation}

\subsubsection{Objective on Candidate Text Perplexity}
As aforementioned, UniTRec leverages perplexity to rank candidate texts. Since lower perplexity indicates higher user-candidate matching probability, regarding the candidate text $Y=[y_{1},y_{2},...,y_{T}]$, we define the perplexity-based matching score $S^{p}$ as its negative perplexity\footnote{Note \href{https://huggingface.co/docs/transformers/perplexity}{https://huggingface.co/docs/transformers/perplexity} for LM perplexity calculation. We empirically discard the outer exponential term in the PPL formula, because it already exists in NCE loss Eq. (\ref{eq4}) and does not affect the final ranking.}:
\begin{equation}\label{eq3}
\setlength{\abovedisplayskip}{8pt}
\setlength{\belowdisplayskip}{8pt}
S^{p} = -\mathrm{PPL}(Y) = \frac{1}{T}\sum\nolimits_{i=1}^{T}\log p_{\theta}(y_{i}|y_{<i})
\end{equation}
, where $p_{\theta}(\cdot)$ denotes the target probability output from the UniTRec Transformer decoder. Similar to Eq.~(\ref{eq2}), we optimize the perplexity-based matching score $S^{p}$ in the NCE loss form. As perplexity empirically varies in a wide range, we introduce a temperature parameter $\tau$ to balance the joint NCE loss gradients following \citet{CLIP}.
\begin{equation}\label{eq4}
\setlength{\abovedisplayskip}{8pt}
\setlength{\belowdisplayskip}{8pt}
\mathcal{L}_{i}^{p} = -\log\frac{\mathrm{exp}(\tau\cdot S^{p+}_i)}{\mathrm{exp}(\tau\cdot S^{p+}_i)+\sum\nolimits_{j=1}^{K}{\mathrm{exp}(\tau\cdot S^{p-}_{j})}}
\end{equation}
, where $\tau$ is learnable and initialized to $1$. On the training dataset $\mathcal{D}$, the joint contrastive learning objective is formulated as:
\begin{equation}
\setlength{\abovedisplayskip}{4pt}
\setlength{\belowdisplayskip}{4pt}
\mathcal{L} = \sum\nolimits_{i=1}^{|\mathcal{D}|}\big(\mathcal{L}_{i}^{d} + \mathcal{L}_{i}^{p}\big)
\end{equation}

\subsection{Model Initialization and Inference}
As UniTRec is a standard text-to-text Transformer, we initialize the parameters from pretrained BART \citep{BART}. In inference, UniTRec predicts the discriminative and perplexity-based scores for each candidate item, respectively. The two separate scores $S^{d}$ and $S^{p}$ are normalized, averaged, and finally ranked as the output. Detailed ranking process is provided in Appendix~\ref{appendix_B}.

\section{Experiments}
We evaluate UniTRec on three text-based recommendation tasks: $1$) \underline{\textit{NewsRec}}, to recommend news articles to users based on their browsing history. We use the \textit{MIND-small} dataset~\citep{MIND} for experiments. 2) \underline{\textit{QuoteRec}}, to recommend quotations to users based on their conversation history. We use the \textit{Reddit-quotation} dataset \cite{quotation_rec1} for experiments. 3) \underline{\textit{EngageRec}}, to recommend social media posts for users to engage with based on their comment history. We use the dataset released by \citet{DCR} for experiments. Detailed dataset statistics is provided in Appendix~\ref{appendix_A}.

\textbf{Implementation Details.} The UniTRec encoder and decoder both consist of $6$ Transformer layers with $768$-dimensional hidden states and $12$ attention heads. We set $L_{1}=3$ and $L_{2}=3$. We use AdamW optimizer~\citep{AdamW} to train UniTRec with cosine learning rate decay.

\begin{table*}[t]
\centering
\vskip -0.74954mm
\resizebox{\textwidth}{!}{
\fontsize{10}{12}\selectfont
\begin{tabular}{l|ccc|ccc|ccc}
\hline
& \multicolumn{3}{c|}{\textit{NewsRec}} & \multicolumn{3}{c|}{\textit{QuoteRec}} & \multicolumn{3}{c}{\textit{EngageRec}} \\
\textbf{Model} & MRR & NDCG@5/10 & HR@5/10 & MRR & NDCG@5/10 & HR@5/10 & MRR & NDCG@5/10 & HR@5/10 \\ \hline
GRU4Rec & 32.91 & 36.20/42.53 & 50.33/68.35 & 34.08 & 34.65/37.93 & 44.45/54.63 & 2.12 & 1.04/1.51 & 1.27/2.65 \\
SASRec & 32.60 & 36.03/42.37 & 50.63/68.64 & 33.63 & 34.30/37.49 & 44.32/54.20 & 2.40 & 1.49/1.95 & 2.16/3.47 \\
BERT4Rec & 32.87 & 36.18/42.40 & 50.21/67.97 & 33.59 & 34.26/37.27 & 43.76/53.05 & 3.04 & 1.98/3.23 & 2.81/6.67 \\
RoBERTa-Sim & 32.96 & 36.47/42.81 & 51.06/69.08 & 37.13 & 37.96/41.18 & 48.14/58.06 & 3.74 & 2.66/3.75 & 4.42/\textbf{7.70} \\
UNBERT & 33.09 & 36.53/42.84 & 50.87/68.82 & 39.75 & 40.74/43.69 & 50.90/60.04 & 2.83 & 1.96/2.67 & 3.11/5.24 \\
\hline
UniTRec & \textbf{33.76} & \textbf{37.63}/\textbf{43.74} & \textbf{52.61}/\textbf{69.89} & \textbf{41.24} & \textbf{42.38}/\textbf{45.31} & \textbf{52.87}/\textbf{61.88} & \textbf{4.06} & \textbf{3.23}/\textbf{4.29} & \textbf{4.58}/7.68 \\
\hline
\end{tabular}
}
\caption{Experiment results on three text-based recommendation tasks. MRR denotes mean reciprocal rank, NDCG denotes normalized discounted cumulative gain, and HR denotes hit ratio (presented in percentage). The overall performance of UniTRec is better than other baseline models with $p$-value $<0.05$, validated by unpaired t-test.}
\label{table:main_experiments}
\end{table*}
\begin{table*}[t]
\centering
\resizebox{\textwidth}{!}{
\fontsize{10}{12}\selectfont
\begin{tabular}{l|ccc|ccc|ccc}
\hline
& \multicolumn{3}{c|}{\textit{NewsRec}} & \multicolumn{3}{c|}{\textit{QuoteRec}} & \multicolumn{3}{c}{\textit{EngageRec}} \\
\textbf{Model} & MRR & NDCG@5/10 & HR@5/10 & MRR & NDCG@5/10 & HR@5/10 & MRR & NDCG@5/10 & HR@5/10 \\ \hline
UniTRec & 33.76 & 37.63/43.74 & 52.61/69.89 & 41.24 & 42.38/45.31 & 52.87/61.88 & 4.06 & 3.23/4.29 & 4.58/7.68 \\
w/o BART Init & 30.31 & 33.32/39.69 & 47.55/65.78 & 19.02 & 17.66/20.80 & 22.45/32.16 & 2.24 & 0.86/1.61 & 1.27/3.62 \\ \hline
w/o Local-Att & 33.34 & 37.22/43.32 & 52.28/69.54 & 40.44 & 41.63/44.56 & 52.09/61.15 & 3.92 & 3.19/4.15 & 4.38/7.36 \\
w/o Global-Att & 33.22 & 37.06/43.17 & 52.14/69.47 & 40.25 & 41.47/44.26 & 52.07/60.76 & 3.64 & 2.78/3.59 & 3.89/6.35 \\ \hline
Disc-Score only & 33.07 & 36.76/43.03 & 51.68/69.46 & 40.59 & 41.81/44.65 & 52.39/61.14 & 3.82 & 2.99/3.60 & 4.49/6.85 \\
PPL-Score only & 32.83 & 36.39/42.59 & 51.05/68.67 & 40.31 & 41.43/44.47 & 52.13/61.20 & 3.29 & 2.39/3.03 & 3.86/5.66 \\
\hline
\end{tabular}
}
\caption{Recommendation performance of ablation model variants.}
\label{table:ablations}
\end{table*}

\textbf{Baselines.} We compare UniTRec with competitive baselines: $1$) \underline{GRU4Rec} \citep{GRU4Rec} utilizes a GRU network to learn multi-turn history. $2$) \underline{SASRec} \citep{SASRec} encodes user history with a self-attention based sequential model. $3$) \underline{BERT4Rec} \citep{BERT4Rec} employs bidirectional self-attention to model user history. $4$) \underline{RoBERTa-Sim}, a simple yet strong baseline mentioned in \citet{Quoter}, uses the hidden states of \textrm{[CLS]} tokens to measure user-candidate similarity. $5$) \underline{UNBERT}, implemented as \citet{UNBERT}, concatenates history and candidate texts as the input to BERT and predicts matching scores from the final hidden states of \textrm{[CLS]} tokens.

Note that we do not consider other methods that use non-text inputs~(e.g., user profile, text topic labels). For fair comparison, all baseline models use pretrained $12$-layer RoBERTa-base \citep{Roberta} as text encoders to learn embeddings of texts.

\subsection{Main Results}
Table~\ref{table:main_experiments} shows the performance of experiment models. From the results of \textit{NewsRec} and \textit{QuoteRec}, we can see that UniTRec outperforms all baseline models by a clear margin. Also, RoBERTa-Sim and UNBERT that directly use the \textrm{[CLS]} hidden states to represent user history, surpass other baselines that build additional aggregation networks upon the whole RoBERTa outputs. As displayed in the results, \textit{EngageRec} is the most difficult task. We inspect the dataset and find that the texts on social media contain too much noise (e.g., URL and emoji), and the user history contains less number of turns. Nevertheless, UniTRec achieves better overall performance than other baseline models, validating its robustness on noisy text inputs and limited user history.

\subsection{Ablation Studies and Analyses}
We further conduct ablation studies on UniTRec. The experiment results are reported in Table~\ref{table:ablations}.

\paragraph{Initialization of UniTRec.} We train \mbox{UniTRec} from scratch without initialization from pretrained BART (refer to {\color{blue}{w/o BART Init}}). The recommendation performance significantly drops in all three tasks, which indicates that acquiring effective text understanding ability from PLM is a necessary key to UniTRec performance.

\paragraph{Local and global attention.} We investigate the function of two-level attention modules of the UniTRec history encoder. Concretely, we set $L_{1}=0$ in {\color{blue}{w/o Local-Att}} and $L_{2}=0$ in {\color{blue}{w/o Global-Att}}, where $L_{1}+L_{2}=6$. We can observe that removing local and global attention from the original UniTRec history encoder both lead to suboptimal performance, while the performance drop is more significant in w/o Global-Att. The results justify the effectiveness of jointly modeling two-level history contexts through adapted Transformer attention masking without additional parameters.

\paragraph{Discriminative and perplexity-based objectives.} We probe into training UniTRec with standalone discriminative ({\color{blue}{Disc-Score only}}) and perplexity-based ({\color{blue}{PPL-Score only}}) contrastive objectives, respectively. We can see that the discriminative objective yields better performance than the perplexity-based objective. Besides, the model performance on both standalone objectives declines compared to the original joint objective. The results indicate that the discriminative and perplexity-based matching scores are complementary and can jointly provide more accurate signals of user history and candidate text matching for text-based recommendation.

\section{Conclusion}
We present a unified Transformer UniTRec for text-based recommendation. UniTRec learns two-level contexts of multi-turn user history and jointly exploits discriminative matching scores and candidate text perplexity as matching objectives. Empirical experiments on three text-based recommendation datasets corroborate the effectiveness of UniTRec.

\section{Limitations}
Our model only focuses on utilizing text information for recommendation, which is a key limitation of this work. In real-world settings, recommender systems are usually required to handle heterogeneous information inputs. UniTRec is a pure text-based recommender modeling user history and candidate texts as inputs. However, incorporating additional side information (e.g., user profile, text topic labels, and dwell time of user behaviors) could further improve the recommendation performance and alleviate the \textit{cold start} problem. Furthermore, UniTRec only models two-level relations of user behavior history. Nonetheless, incorporating more user behavior information, such as implicit and negative feedback, could further enhance the recommendation performance.

\section*{Acknowledgements}
We appreciate constructive comments from anonymous reviewers. The research described in this paper is partially supported by CUHK under Project No. 3230366.

\bibliography{anthology,custom}
\bibliographystyle{acl_natbib}
\clearpage
\appendix
\begin{table}[t]
\centering
\resizebox{80mm}{!}{
\fontsize{10}{12}\selectfont
\begin{tabular}{l|c|c|c}
\hline
Dataset & \textit{NewsRec} & \textit{QuoteRec} & \textit{EngageRec} \\ \hline
Avg. history turns & 26.09 & 4.24 & 3.29 \\
Avg. history tokens & 414.40 & 279.82 & 286.82 \\
\hline
Avg. candidates & 37.23 & 1111 & 7163 \\
Avg. candidate tokens & 16.15 & 19.11 & 102.42 \\
\hline
\end{tabular}
}
\captionsetup{font=10pt}
\caption{Statistics of three text-based recommendation training datasets. History and candidate \underline{\textbf{tokens}} denote the number of BPE-tokenized tokens. The test set distribution is closed to the training sets (except candidates of \textit{EngageRec}) and hence omitted. Note that the max length of each history log is truncated to $1024$ tokens.}
\label{table:datasets}
\end{table}
\section{Dataset Statistics}\label{appendix_A}
The detailed statistics of the three text-based recommendation datasets are displayed in Table \ref{table:datasets}. Note that we use news titles as the text inputs for \textit{NewsRec} following \citet{HieRec}. \textit{NewsRec} regards the user clicked and non-clicked news as candidate texts, while \textit{QuoteRec} and \textit{EngageRec} regard all potential quotation texts and post texts as candidates. Different from \citet{DCR} that formulates the task as recommending candidate users to given posts based on post content, we formulate the task as recommending candidate posts to given users based on user history.

\begin{algorithm}[h]
\caption{Candidate Ranking Processs}
\label{alg:rank2}
\begin{algorithmic}[1]
\algsetup{linenosize=\small}
\small
\REQUIRE discriminative scores $S^d=\{S^d_1,S^d_2,...,S^d_M\}$, \\
         \quad\ perplexity-based scores $S^p=\{S^p_1,S^p_2,...,S^p_M\}$.
\ENSURE final averaged ranking $\bar{R}$.

\STATE Derive the normalized discriminative scores $S^d_{norm}=\mathrm{softmax}(S^d)$.

\STATE Derive the normalized perplexity-based scores $S^p_{norm}=\mathrm{softmax}(S^p)$.

\STATE Derive the geometric average scores $\bar{S}=\log{(S^d_{norm})}+\log{(S^p_{norm})}$.

\STATE Sort the averaged scores $\bar{S}$ by descending order to derive the final ranking: $\bar{R}\gets\mathrm{Rank_{des}}(\bar{S})$.

\RETURN $\bar{R}$

\end{algorithmic}
\end{algorithm}
\section{Inference Ranking}\label{appendix_B}
Given the user history and $M$ candidate texts, UniTRec first predicts the discriminative ranking scores $S^d=\{S^d_1,S^d_2,...,S^d_M\}$ and perplexity-based ranking scores $S^p=\{S^p_1,S^p_2,...,S^p_M\}$ of the candidates. Algorithm \ref{alg:rank2} outlines an approach to aggregate the final ranking based on $S^d$ and $S^p$. Note that the function $\mathrm{Rank}(S)$\footnote{$\mathrm{Rank}(S)$ works similarly to \href{https://docs.scipy.org/doc/scipy/reference/generated/scipy.stats.rankdata.html}{\textrm{scipy.stats.rankdata}()}. For example in ascending order, $\mathrm{Rank_{asc}}(\{0.2, 0.6, 0.7, 0.4\})=\mathrm{scipy.stats.rankdata}([0.2, 0.6, 0.7, 0.4])=[1,3,4,2]$} denotes outputting the sorted order of elements in a score list $S$. There exist other ways to average the ranking of $S^d$ and $S^p$, which we leave for future work to explore.

\section{Qualitative Analysis}\label{appendix_C}
We show randomly sampled outputs of UniTRec, for instance, demonstrated on the news recommendation and quote recommendation tasks. Table \ref{table:visualization1} and \ref{table:visualization2} showcase the qualitative samples.
\begin{table*}[t]
\centering
\begin{subtable}
\centering
\fontsize{7.5}{9.5}\selectfont
\begin{tabularx}{\textwidth}{l|X}
\Xhline{1pt}
Turn& {\textit{\qquad\qquad\qquad\qquad\qquad\qquad\qquad\qquad\qquad\qquad\qquad\qquad History News Texts}} \\ \hline
$\#1$ & Mac Engel: As long as these results are acceptable, Dallas Cowboys will continue to be losers \\
$\#2$ & NFL world reacts to officials handing Packers win over Lions \\
$\#3$ & Maryland Congressman Elijah Cummings, a Democrat and Chair of House Oversight and Reform Committee, has died: CNN \\
$\#4$ & Unprecedented movement detected on California earthquake fault capable of 8.0 temblor \\
$\#5$ & Bag Explodes While Being Loaded On Volaris Flight At Midway Airport \\
$\#6$ & Orlando Scandrick rips Eagles: They have "accountability issues" \\
$\#7$ & Meghan King Edmonds, Jim Edmonds' Nanny Denies Cheating Allegations \\
$\#8$ & Nearly \$400M worth of cocaine and marijuana intercepted by US Coast Guard \\
$\#9$ & Former NBA first-round pick arrested in sex sting operation \\
$\#10$ & China's trade with US shrinks in October despite optimism \\
\Xhline{1pt}
\end{tabularx}
\begin{tabular}{ccccc}
\\
\Xhline{1pt}
{\textit{Candidate News Texts}} & $S^d$ & $S^p$ & $\bar{R}$ & Clicked\\ \hline
Taylor Swift Rep Hits Back at Big Machine, Claims She's Actually Owed \$7.9 Million in Unpaid Royalties &$0.095$&$0.069$&$4$&\xmark \\
Former North Carolina State, NBA player Anthony Grundy dies in stabbing, police say &$0.172$&$0.155$&$3$&\xmark \\
13 Reasons Why's Christian Navarro Slams Disney for Casting "the White Guy" in The Little Mermaid &$0.048$&$0.065$&$7$&\xmark \\
Opinion: Colin Kaepernick is about to get what he deserves: a chance &$0.303$&$0.250$&$1$&\checkmark \\
3 Indiana judges suspended after a night of drinking turned into a White Castle brawl &$0.076$&$0.059$&$5$&\xmark \\
66 Cool Tech Gifts Anyone Would Be Thrilled to Receive &$0.009$&$0.005$&$9$&\xmark \\
Police find 26 children behind false wall at Colorado day care &$0.034$&$0.116$&$6$&\xmark \\
I've been writing about tiny homes for a year and spent 2 nights in a 300-foot home to see what it is all about &$0.029$&$0.019$&$8$&\xmark \\
Report: Police investigating woman's death after Redskins' player Montae Nicholson took her to hospital &$0.235$&$0.261$&$2$&\checkmark \\
\Xhline{1pt}
\end{tabular}
\vskip -1.5mm
\caption*{(\romannumeral 1) Qualitative Example-A from news recommendation.}
\vskip -2mm
\end{subtable}

\begin{subtable}
\centering
\fontsize{7.5}{9.5}\selectfont
\\
\begin{tabularx}{\textwidth}{l|X}
\Xhline{1pt}
Turn& {\textit{\qquad\qquad\qquad\qquad\qquad\qquad\qquad\qquad\qquad\qquad\qquad\qquad History News Texts}} \\ \hline
$\#1$ & Toddler dancing to celebrate 11 months cancer-free goes viral \\
$\#2$ & NFL Week 8 Power Rankings: Old-school football rules the day \\
$\#3$ & The 25 US cities where it's easiest to get a mortgage \\
$\#4$ & Burning questions for Cowboys vs Giants on "Monday Night Football" \\
$\#5$ & Who's the favorite to win 2019 NFL rushing title? \\
$\#6$ & Grading all 32 NFL teams heading into the last eight weeks of the 2019 season \\
$\#7$ & Jennifer Aniston looks amazing in a makeup-free selfie, plus more news \\
$\#8$ & This \$12 million "mansion yacht" is made entirely of stainless steel and it's a first for the industry. Take a peek inside \\
\Xhline{1pt}
\end{tabularx}
\begin{tabular}{ccccc}
\\
\Xhline{1pt}
{\textit{Candidate News Texts}} & $S^d$ & $S^p$ & $\bar{R}$ & Clicked\\ \hline
Opinion: Colin Kaepernick is about to get what he deserves: a chance &$0.330$&$0.400$&$1$&\checkmark \\
U.S. Troops Will Die If They Remain in Syria, Bashar Al-Assad Warns &$0.024$&$0.011$&$10$&\xmark \\
Pete Davidson, Kaia Gerber Are Dating, Trying to Stay "Low Profile" &$0.064$&$0.033$&$6$&\xmark \\
The Hottest Tech Gifts This Holiday Season &$0.050$&$0.027$&$8$&\xmark \\
Taylor Swift Rep Hits Back at Big Machine, Claims She's Actually Owed \$7.9 Million in Unpaid Royalties &$0.046$&$0.038$&$7$&\xmark \\
13 Reasons Why's Christian Navarro Slams Disney for Casting "the White Guy" in The Little Mermaid &$0.060$&$0.096$&$4$&\checkmark \\
Some believe Mason Rudolph, hit in head with his own helmet, isn't getting enough blame &$0.154$&$0.179$&$2$&\checkmark \\
South Carolina teen gets life in prison for deadly elementary school shooting &$0.066$&$0.046$&$5$&\xmark \\
The Unlikely Star of My Family's Thanksgiving Table &$0.047$&$0.021$&$9$&\xmark \\
Police investigating woman's death after Redskins' player Montae Nicholson took her to hospital &$0.158$&$0.149$&$3$&\xmark \\
\Xhline{1pt}
\end{tabular}
\vskip -1.5mm
\caption*{(\romannumeral 2) Qualitative Example-B from news recommendation.}
\end{subtable}
\vskip 3mm
\caption{Case analyses of news recommendation. \textit{History News Texts} are sorted by user-clicked timestamps. $S^d$, $S^p$, and $\bar{R}$ are normalized discriminative, perplexity-based scores, and average ranking as described in Appendix~\ref{appendix_B}. \underline{\textbf{Clicked}} denotes the ground truth user-click labels. Note that the experiment history logs are anonymized and delinked, which is always the first priority of the recommendation study.}
\label{table:visualization1}
\end{table*}
\begin{table*}[t]
\centering
\begin{subtable}
\centering
\fontsize{7.5}{9.5}\selectfont
\begin{tabularx}{\textwidth}{l|X}
\Xhline{1pt}
Turn& {\textit{\qquad\qquad\qquad\qquad\qquad\qquad\qquad\qquad\qquad\qquad\quad\, Conversation Threading History}} \\ \hline
$\#1$ & I own an FJ. It's a great car and even on stockies. It s great offroad. \\
$\#2$ & I feel bad for you that you run the risk of being associated with the typical FJ owner. \\
$\#3$ & What is a typical FJ owner? I've not heard anything bad about FJ owners. \\
$\#4$ & It's like someone who drives a jeep wrangler in NYC. There's no need. Tons of FJ owners do that have it and not use it for what it's made for. \\
$\#5$ & God forbid someone likes the design of a car and doesn't use it offroad. \\
$\#6$ & Then buy a much more economic environmentalist friendly version. If you buy something and always use it for much less than it's purpose, why buy it? \\
$\#7$ & Or people can buy whatever the hell they want because it's their money and not yours. \\
$\#8$ & You're entirely right. Just like people can be rude just because you can do it, because you have the ability but why should you ass. \\
$\#9$ & I wasn't aware that somebody buying a vehicle that they like and you don't was morally wrong. \\
$\#10$ & I love FJs. It's perfectly fine to buy whatever you think looks nice. \\
\Xhline{1pt}
\end{tabularx}
\begin{tabular}{ccccc}
\\
\Xhline{1pt}
{\textit{Candidate Quote Texts}} & $S^d$ & $S^p$ & $\bar{R}$ & Ground truth\\ \hline
Beauty is in the eye of the beholder.&$0.480$&$0.471$&$1$&\checkmark \\
A fool and his money are soon parted.&$0.176$&$0.140$&$2$& \\
Form follows function.&$0.051$&$0.046$&$3$& \\
Everything is worth what its purchaser will pay for it.&$0.040$&$0.058$&$4$& \\
Because it's there.&$0.038$&$0.029$&$5$& \\
You can't fix stupid.&$0.021$&$0.034$&$6$& \\
The lady doth protest too much, methinks.&$0.022$&$0.013$&$7$& \\
It's all about the money.&$0.020$&$0.013$&$8$& \\
Anybody driving slower than you is an idiot, and anyone going faster than you is a maniac?&$0.012$&$0.018$&$9$& \\
Opportunity is missed by most people.&$0.018$&$0.008$&$10$& \\
\Xhline{1pt}
\end{tabular}
\vskip -1.5mm
\caption*{(\romannumeral 3) Qualitative Example-C from quote recommendation.}
\vskip -2mm
\end{subtable}

\begin{subtable}
\centering
\fontsize{7.5}{9.5}\selectfont
\\
\begin{tabularx}{\textwidth}{l|X}
\Xhline{1pt}
Turn& {\textit{\qquad\qquad\qquad\qquad\qquad\qquad\qquad\qquad\qquad\qquad\quad\, Conversation Threading History}} \\ \hline
$\#1$ & Society is becoming more efficient, which is a good thing. People should realize there's no point in holding back this technology just for the sake of keeping people employed. If this were beneficial, then calculators and computers shouldn't exist either. \\
$\#2$ & One small problem is that people need to pay rent and eat. \\
$\#3$ & So we should ditch computers and go back to the typing pool? Should we get rid of heavy earth moving equipment and just use hundreds of guys with hand tools to build everything? It would employ a hell of a lot more people. \\
$\#4$ & No one's saying that. I don't think anyone is really against automation, but as it increases, there are soon going to be more people that there are jobs that actually need doing. I actually believe we've already passed this point. So what do we do with the people, who can't get jobs simply because there are none? It's an issue that need assessed immediately. \\
$\#5$ & Tons and tons and tons of American jobs have been replaced by new jobs created by technology or in support of technology years ago. An office might have needed people to handle filing paperwork, keeping it in order, and retrieving, where now a document management system has made them completely redundant. The upshot is that to access that DMS, people are out there selling computers, installing computers, servicing computers, and supporting end users building the servers installing, supporting monitoring backing them up, and all that jobs that come in support of those progress is progress. And it advances human efficiency and knowledge. These are just one or two examples, but the answer is not to kill progress. Other countries simply won't. The answer is to push education to the forefront, so people are prepared for these jobs and whatever other challenges the future may bring. \\
$\#6$ & This is true. But it s unfortunate technological advances tend to reduce low skill jobs and replace them with high skill jobs. It would feel more fair if the low skilled workers could all do training programs and become high skilled workers. But this isn't really the case. Those jobs end up being taken by someone who had better educational opportunities or someone younger who still has time to take advantage of education. \\
$\#7$ & The reality is the reality. Unfortunate or not educating people will create more educated people to handle high skill jobs, and I'll tell you being a desktop support technician isn't high skill. As that's where we push in the future, any amount of hand wringing won't change the facts. We must educate our people if we want to be a global leader in more than homelessness poverty. \\
$\#8$ & Education won't matter. We are at the end of the job age at some point in the near future. We are going to have to deal with the fact that getting a job isn't a reality for a significant percentage of the population. Society will have to radically change as it did during the industrial revolution. \\
$\#9$ & Much cheaper to heavily discourage having more children free abortions. Then in years there won't be so many useless people who can apparently be replaced by a simple robot. \\
$\#10$ & Virtually every job will be replaced by automation name skilled trades that can't be automated. I imagine you'd be surprised at how hard this is. Are pharmacists useless, surgeons, accountants? I'd bet that your job is just as replaceable as these. \\
\Xhline{1pt}
\end{tabularx}
\begin{tabular}{ccccc}
\\
\Xhline{1pt}
{\textit{Candidate Quote Texts}} & $S^d$ & $S^p$ & $\bar{R}$ & Ground truth\\ \hline
There's no such thing as a free lunch.&$0.365$&$0.417$&$1$& \\
I can't predict the future.&$0.185$&$0.210$&$2$&\checkmark \\
I have never let my schooling interfere with my education.&$0.104$&$0.059$&$3$& \\
Prevention is better than cure.&$0.044$&$0.083$&$4$& \\
Knowledge is power.&$0.059$&$0.052$&$5$& \\
Don't let schooling interfere with your education.&$0.044$&$0.043$&$6$& \\
Nature abhors a vacuum.&$0.036$&$0.024$&$7$& \\
There is no substitute for hard work.&$0.024$&$0.017$&$8$& \\
There are three kinds of lies: lies, damned lies, and statistics.&$0.022$&$0.013$&$9$& \\
You can't fix stupid.&$0.019$&$0.010$&$10$& \\
\Xhline{1pt}
\end{tabular}
\vskip -1.5mm
\caption*{(\romannumeral 4) Qualitative Example-D from quote recommendation.}
\end{subtable}
\vskip 3mm
\caption{Case analyses of quote recommendation. We demonstrate the candidate quotes of the top $10$ rankings out of all candidates. Note that there is only one ground truth quote for each conversation history.}
\label{table:visualization2}
\end{table*}

\label{sec:appendix}
\end{document}